\documentclass{article}
\usepackage{graphicx}

\usepackage[preprint]{neurips}

\usepackage[utf8]{inputenc} % allow utf-8 input
\usepackage[T1]{fontenc}    % use 8-bit T1 fonts
\usepackage{hyperref}       % hyperlinks
\usepackage{url}            % simple URL typesetting
\usepackage{booktabs}       % professional-quality tables
\usepackage{amsfonts}       % blackboard math symbols
\usepackage{nicefrac}       % compact symbols for 1/2, etc.
\usepackage{microtype}      % microtypography
\usepackage{xcolor}         % colors
\usepackage{ulem}
\usepackage{natbib}
\usepackage{multirow}
\usepackage{subfig}
\usepackage{float} 
\usepackage{algorithm}      % 伪代码环境
\usepackage{algpseudocode}  % 伪代码语法包
\usepackage{amsmath}
\usepackage{graphicx}
\usepackage{amsmath}
\usepackage{amssymb}

\title{RAIDEN-R1: Improving Role-awareness of LLMs via GRPO with Verifiable Reward} 
\author{
Zongsheng Wang$^{1}$, Kaili Sun$^{1}$, Bowen Wu$^{2,1}$, Qun Yu$^{1}$, Ying Li$^{2}$, Baoxun Wang$^{1}$ \\
$^1$Platform and Content Group, Tencent \\
$^2$School of Software \& Microelectronics, Peking University, Beijing, China \\
{\tt \{jasoawang, kailisun, sparkyu, asulewang\}@tencent.com} \\
{\tt{\{jason\_wbw,li.ying\}@pku.edu.cn}}\\
}

\begin{document}

\maketitle

\begin{abstract}
Role-playing conversational agents (RPCAs) face persistent challenges in maintaining role consistency. 
To address this, we propose \textbf{RAIDEN-R1}, a novel reinforcement learning framework that integrates Verifiable Role-Awareness Reward (\textbf{VRAR}). 
The method introduces both singular and multi-term mining strategies to generate quantifiable rewards by assessing role-specific keys.
Additionally, we construct a high-quality, role-aware Chain-of-Thought dataset through multi-LLM collaboration, and implement experiments to enhance reasoning coherence. 
Experiments on the RAIDEN benchmark demonstrate RAIDEN-R1’s superiority: our 14B-GRPO model achieves 88.04\% and 88.65\% accuracy on Script-Based Knowledge and Conversation Memory metrics, respectively, outperforming baseline models while maintaining robustness. 
Case analyses further reveal the model’s enhanced ability to resolve conflicting contextual cues and sustain first-person narrative consistency. 
This work bridges the non-quantifiability gap in RPCA training and provides insights into role-aware reasoning patterns, advancing the development of RPCAs.

\end{abstract}

% \paragraph{Keywords: Role-Playing Conversational Agents, Reinforcement Learning, Quantifiable Reward, Chain-of-Thought Reasoning, Role-Awareness}

\begin{figure*}[!htb]
    \centering
    \includegraphics[width=1.\linewidth]{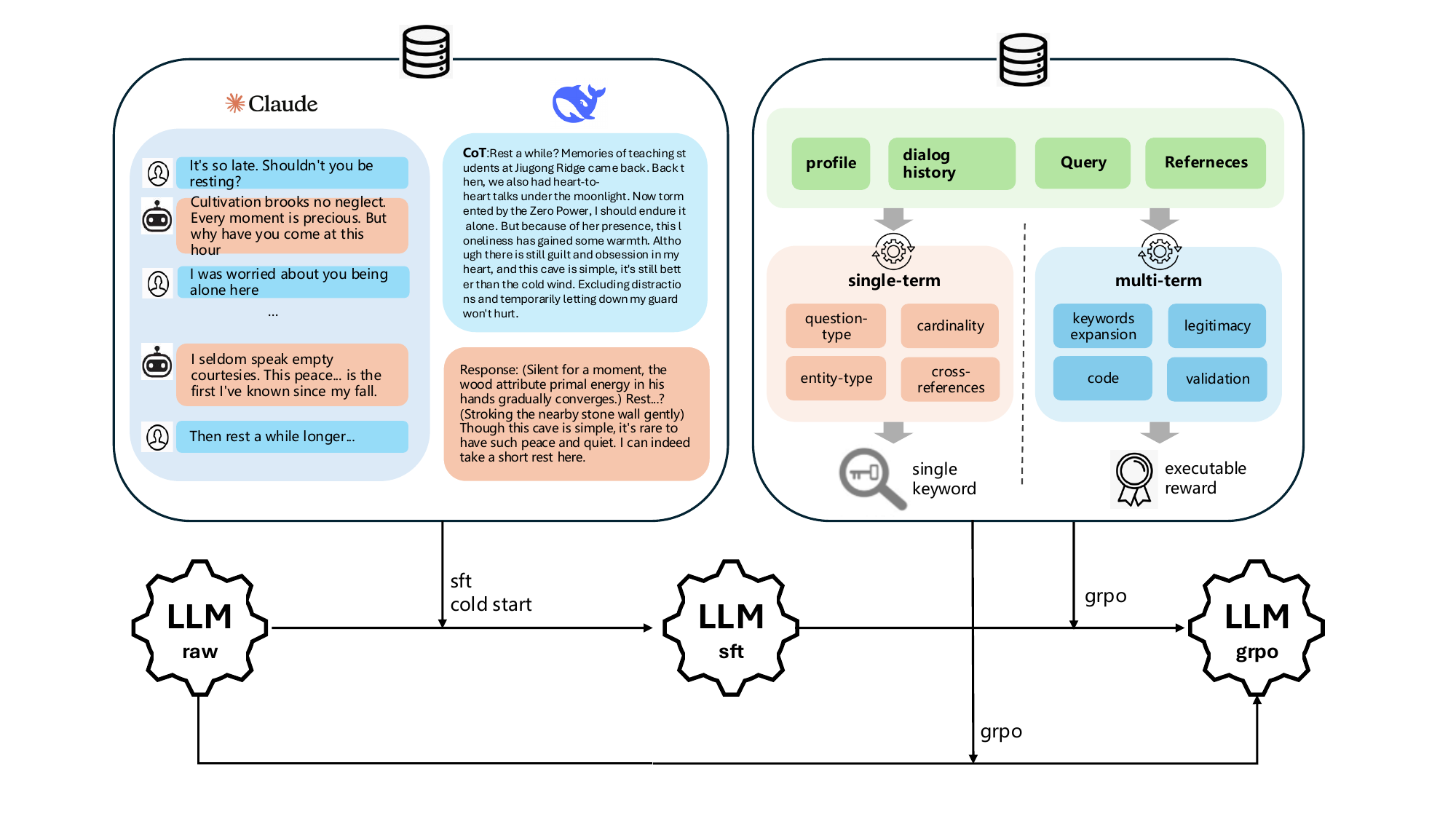}
    \caption{Workflow of \textbf{VRAR}, followed by the multi-stage training strategy of DeepSeek R1, we create two datasets for sft and reinforcement learning.}
    \label{fig:workflow}
\end{figure*}

\section{Introduction}

With recent advances in Large Language Models (LLM)\cite{hurst2024gpt, achiam2023gpt}, Role-Playing Conversational Agents (RPCAs) have become a cutting-edge in AI research. 
Such agents are designed to simulate distinct personas, ranging from fictional characters to celebrities, while maintaining role-aware cognition on their predefined profiles. 
Industrial implementations such as Character.ai and Talkie exemplify this trend, using customized character creation platforms to attract millions of daily active users.
Meanwhile, existing studies such as CharacterGLM\cite{Characterglm} attempt to improve the performance of RPCAs, mainly through data-driven strategies such as synthesizing a higher quality dialogue corpus~\cite{lu2024large,yu2024beyond,wang2024rolellm}. 
To better evaluate RPCAs, benchmarks like CharacterEval\cite{tu-etal-2024-charactereval}, Roleinteract\cite{chen2024roleinteract}, Raiden\cite{wu-etal-2025-raiden}, etc., are also proposed to quantify the agents' performances in terms of self-awareness and conversational abilities~\cite{zhou2025characterbench,chen2024socialbench}. 

Currently, supervised fine-tuning (SFT) remains the dominant paradigm for training RPCAs, where agents are trained to generate persona-specific responses directly without any intermediate reasoning step.
Therefore, conventional RPCAs still suffer from the role-drift problem, attributed to their lack of intrinsic reasoning mechanisms to alleviate conflicting contextual cues.
Some recent research progress is expected to be valuable for handling such problems,
and especially, the studies on the chain-of-thought (CoT) based reasoning have shown the great potential\cite{wei2022chain, zhang2025cot, wang2022iteratively}.
OpenAI-O1\cite{openai2024openaio1card} first implements the complicated CoT for training and achieves unprecedented performance gains over conventional architectures. 
This trend may be further stimulated by the noteworthy work like DeepSeek-R1\cite{guo2025deepseek}, 
which activates self-reasoning capability through Group Relative Policy Optimization (GRPO)\cite{shao2024deepseekmath} with quantifiable rewards. 

Intuitively, the CoT reasoning exhibits natural compatibility with RPCAs, as its step-by-step reasoning mirrors the human-like narrative construction. 
However, a recent study\cite{feng2025reasoning} reveals that, given a general-purpose model without specialized role-playing reinforcement learning, its CoT capacity may even have a detrimental effect on its role-playing performance. 
Even though prior work like DeepSeek-R1 has demonstrated success in applying GRPO on tasks with unique correctness criteria, extending these approaches to role-playing encounters a challenge of non-quantifiability: 
it is difficult to find a unique solution,
since any response consistent with the context and profile could be considered appropriate. 
Whereas employing LLMs to give rewards in GRPO may appear feasible, this approach suffers from subjective bias as it fails to provide clear and actionable gradients for model optimization.
Consequently, designing quantifiable yet effective reward for RPCAs, so
as to improve their role-aware capabilities, represents a critical and unresolved challenge.

In this paper, we propose RAIDEN-R1: a reinforcement learning framework for training RPCAs with efficient and effective role-aware reasoning, based on the RAIDEN benchmark\cite{wu-etal-2025-raiden}. 
% To address the non-quantifiability challenge in role-playing, we attempt to solve it from scratch: constructing simple but accurate rule-based rewards inspired by self(role)-awareness metrics in Raiden benchmarks\footnote{other metrics related to conversational abilities will be included in our future works}. 
We attempt to address the non-quantifiability challenge from scratch,
by presenting \textbf{VRAR}~\ref{fig:workflow}: a simple and \textbf{V}erifiable rule-based method inspired by \textbf{R}ole-\textbf{A}wareness metrics in Raiden benchmarks\footnote{Other metrics related to conversational abilities will be included in our future works.} as \textbf{R}eward. 
% Specifically, we develop an automated pipeline that first collects role-aware related questions which are prone to making mistakes, then extracts rule-compatible rewards through a combination of heuristic rules and LLM-based mining.
In detail, we have proposed the rule-compatible rewards on the basis of the role-awareness metrics in Raiden benchmarks, 
and correspondingly, developed a pipeline for automatically collecting the role-aware error-prone questions.
Our best implementation of GRPO on extracted data with quantifiable rewards demonstrates measurable improvements in role-aware metrics and superior robustness over SFT.

% During experiments, we have also observed instability in the training process. 
% To address this challenge, we systematically investigate multiple variant designs of format and accuracy rewards on different scales of base models.

Furthermore, we also implement cold-start experiments to replicate the multi-stage training strategy of DeepSeek-R1. 
The construction pipeline of cold-start corpus involves interaction among multi-LLMs to generate higher-quality first-person CoT reasoning and corresponding responses without human annotations. 
This approach enables smaller-scale models to rapidly learn role-aware CoT reasoning with consistent first-person narrative. 
% while achieving notable improvements in conversational performance.
Finally, we conduct an in-depth case analysis that reveals several critical findings regarding our model's reasoning patterns. 
Specifically, we examine how our model thinks and constructs responses under normal dialogue contexts or misleading queries while maintaining role consistency.

Overall, our contributions are highlighted as follows: 
\begin{itemize}
\item We propose a simple yet effective reward design methodology with the corresponding data generation pipeline to address the non-quantifiability challenge in RPCAs, which proves to be effective on Role-Playing Conversational Agents.
% \item A high-quality CoT dataset specifically designed for role-playing SFT is constructed to replicate the multi-stage training strategy of DeepSeek-R1. 
% Such approach enables model to think longer and more informative and achieve higher scores in role-play related evaluations.
\item A high-quality COT corpus specifically designed for role-playing SFT has been meticulously constructed to replicate the multi-stage training strategy of DeepSeek-R1. This approach enables the model to engage in extended, more informative reasoning processes, thereby achieving superior performance metrics in role-play evaluation paradigms.
\item We present intriguing findings, particularly regarding how the model thinks and reasons in different role-play scenarios, which offer insights for future research methodologies. 

\end{itemize}

\section{Methodology}

% 使用 minipage 分左右两栏
\noindent
\begin{figure}[t]
  \centering
    \begin{minipage}[t]{0.57\textwidth}  % 左栏（伪代码）
      \begin{algorithm}[H]
        \caption{Accuracy Reward}
        \label{alg:accuracy reward}
        \begin{algorithmic}[1]
            \State \textbf{Input:} A triplet $ (L, X, R) $, where:
                    \State \quad $ L $: Class label $ L \in \{\text{STV}, \text{MTDP}\} $
                    \State \quad $ X $: Keyword $ k $ if $ L = \text{STV} $, else parsing functions $ f $
                    \State \quad $ R $: Response from the model
            \State \textbf{Output:} Accuracy Reward Score $ S \in \{0, 1\} $
            \If{$ L = \text{STV} \ \mathbf{and} \ X \in R $} 
                \State $ S \gets 1 $
            \ElsIf{$ L = \text{MTDP} \ \mathbf{and} \ X(R) = \text{True} $} 
                \State $ S \gets 1 $
            \Else 
                \State $ S \gets 0 $
            \EndIf
        \end{algorithmic}
      \end{algorithm}
    \end{minipage}
    \hfill  % 填充水平间距
    \begin{minipage}[t]{0.41\textwidth}  % 右栏（图片）
      \begin{figure}[H]
        \centering
        \includegraphics[width=\linewidth]{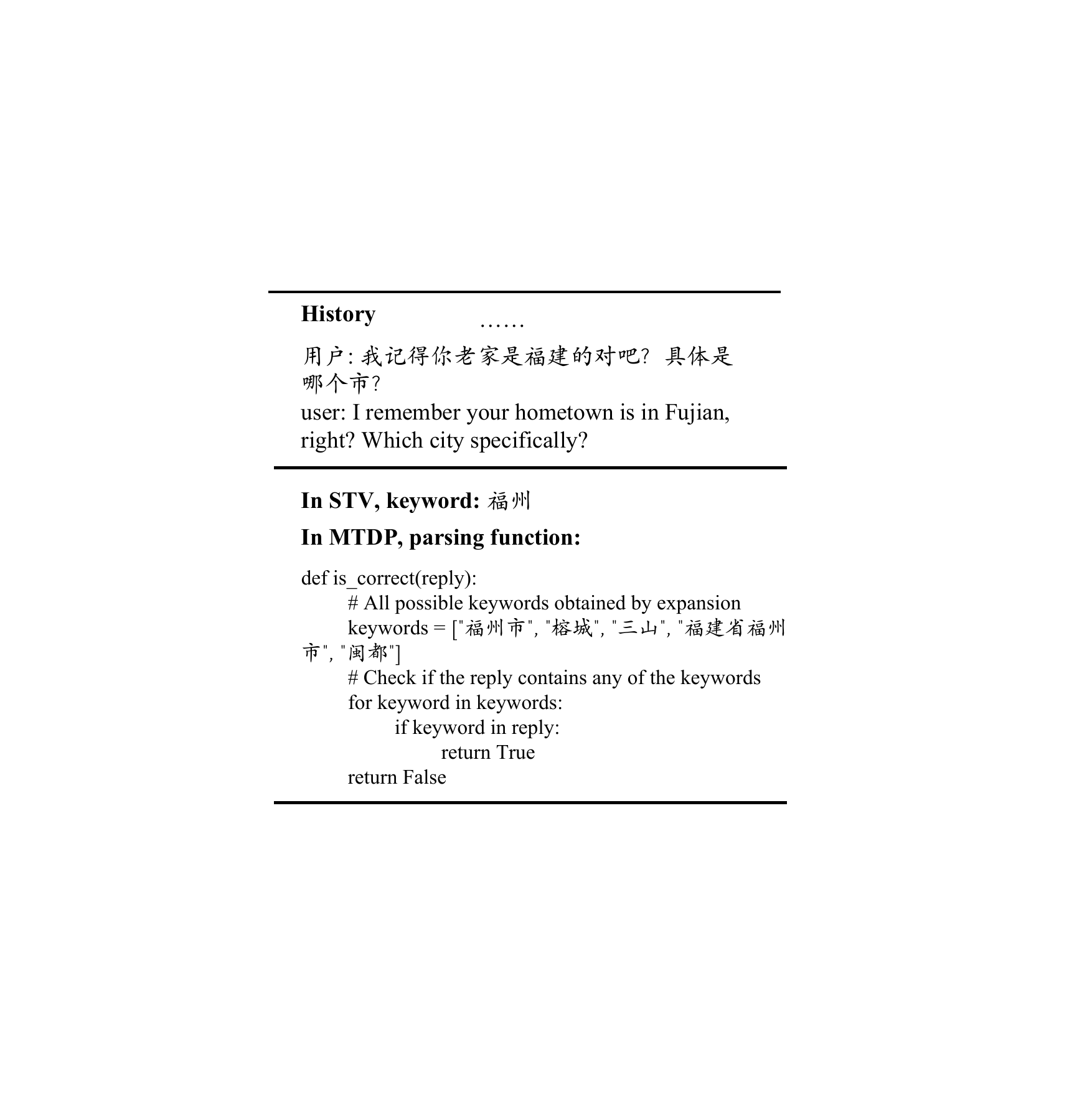}  % 替换为你的图片路径
        \label{fig:example}
      \end{figure}
    \end{minipage}
\caption{Pseudocode and example of one parsing function in accuracy reward.}
\end{figure}

%% to be added: raiden数据，以及各个评测维度的介绍，用于解释角色扮演对话的指标
% Evaluating a role-playing dialogue system typically involves multiple dimensions such as character-consistency, conversational memory and linguistic style, which are inherently challenging to quantify into an objective metric. Whereas employing LLMs as evaluators may appear feasible, this approach suffers from inefficiency and a critical limitation: LLM-based assessments tend to be overly subjective. Such subjectivity leads to significant challenges for optimization methods like GRPO, as it fails to provide clear, actionable gradients for model improvement.

% Consequently, we attempt to solve this problem from scratch: exploiting two relatively objective role-playing dimensions - character-consistency and conversational memory, and extract keywords from dialogue corpus that deterministically verify response correctness. Other dimensions such as linguistic style and more advanced reasoning skill will be included in our future works.

\subsection{Reinforcement Learning}

During the reinforcement learning phase, we focus on enhancing the role-aware capabilities of role-playing models, by conducting the DeepSeek-R1-style training. To address the challenge of quantitatively evaluating role-playing responses, we propose the \textbf{VRAR} framework which implements a keyword-driven reward mechanism designed to systematically assess role consistency and contextual alignment in generated responses. 
On this basis, we develop two distinct data production workflows: Single-Term Validation for granular keyword verification and Multi-Term Dynamic Parsing, implementing Python verification functions with recursive parsing frameworks. 
The former approach addresses scenarios with accurate responses having limited variations, while the latter manages cases where responses are more diverse yet remain controllable through systematic keyword cooperations. 
The \textbf{VRAR} framework enables measurable evaluation metrics while maintaining adaptability to diverse role-playing scenarios, offering a structured methodology for optimizing role-aware behavioral patterns in language models.

\subsubsection{Data Collecting}

Our data sources are divided into two components: the RAIDEN benchmark and a general role-playing dataset. 
The RAIDEN benchmark with a measurement-driven custom conversational dataset is specifically designed for evaluating RPCAs, where each dialogue turn contains explicitly annotated evaluation objectives and reference responses\cite{wu-etal-2025-raiden}. 
To enhance the role-awareness of RPCAs as targeted in this study, we selected data from the Script-Based Knowledge (SBK) and Conversation Memory (CM) dimensions for training, due to their definitive answer characteristics. 
For the general role-playing dataset lacking predefined evaluation metrics, we generated questions based on character profiles and dialogue histories.
Furthermore, we employed the Qwen2.5-14B-Instruct~\cite{yang2024qwen2} model to filter questions, retaining incorrectly answered ones as challenging data, while supplementing with minimal simple data adjusted from perspectives of dialogue history length and question types to optimize data distribution.

Subsequently, we employed Claude 3.5~\footnote{https://www.anthropic.com/} to extract keywords from corresponding responses for each question. 
Considering the data diversity and varying difficulty levels, we accordingly designed two distinct data production workflows: Single-Term Validation and Multi-Term Dynamic Parsing.

% Considering the diversity of data and varying difficulty levels, we designed two distinct data production workflows: Single-Term Validation (STV) and Multi-Term Dynamic Parsing (MTDP).

\paragraph{Single-Term Validation}~{}
\newline
% \subsubsection{Single-Term Validation}
% to be added: 配图解释
% Building upon the Raiden benchmark dataset, we design an automated pipeline focusing on 'Script-Based Knowledge' and 'Conversation Memory dimensions'. Such system employs Claude 3.5 as the primary LLM judge, enforcing the following criteria:
Single-Term Validation applies to scenarios with explicit and unique keywords, where data filtering adheres to the following criteria:
\begin{itemize}
\item \textbf{Question-Type Filtering}: Only WH-questions undergo keyword extraction, while polar ("Is this...?") and alternative ("Is this or isn't this...?") questions are excluded.
\item \textbf{Entity-Type Validation}: Excluding samples if either no unambiguous keywords can be found, or extracted terms are non-nominal and fail to represent distinct entities.
\item \textbf{Cardinality Constraint}: Strictly retain only samples containing exactly one validated keyword.
\item \textbf{Multi-References Verification}: For each dialogue history, generate reference outputs from multiple models as references (GPT-4, MiniMax-abab6-chat, Baichuan-NPC\cite{yang2023baichuan}, GPT-3.5). A keyword is considered valid only if it appears consistently across all references.
\end{itemize}

Such an approach maximizes the precision of single-keyword extraction. When a response contains the validated keyword, we assert it with high-confidence correctness.

% \subsubsection{Multi-Term Dynamic Parsing}
\paragraph{Multi-Term Dynamic Parsing}~{}  
\newline
Building upon scenarios requiring dynamic semantic equivalence across multiple keywords, the Multi-Term Dynamic Parsing method validates responses against contextual and role-aware criteria using:
\begin{itemize}
\item \textbf{Keyword Expansion}: Utilize QwQ-32B~\cite{qwq32b} to expand existing keywords into variant keywords with equivalent semantic meanings.
\item \textbf{Legitimacy Verification}: Employ Qwen-72B to evaluate the relevance of expanded keywords and filter out irrelevant terms.
\item \textbf{Python Code Generation}: Generate a Python code snippet using QwQ-32B based on expanded keywords to assess the correctness of responses. 
\item \textbf{Validation}: Evaluate the correctness of ten existing model responses using both QwQ-32B judgments and corresponding Python code executions. Retain data instances where determination results show over 70\% consistency between both validation methods.
\end{itemize}

\begin{algorithm}[t]
    \caption{Format Reward}
    \label{alg:forward reward}
    \begin{algorithmic}[1]
        \State \textbf{Input:} Response from the model $ R $ 
        \State \textbf{Constant:} Special vocabulary $ W = \{w_1, w_2, \dots, w_n\} $
        \State \textbf{Output:} Format Reward Score $ S \in \{0, 1\} $
        \If{ $
          \Call{CheckPattern}{R} 
          \ \mathbf{and} \ 
          \Call{CalculateChineseRatio}{R} > 0.7 
          \ \mathbf{and} \ 
            R \notin W 
          $}
            \State $ S \gets 1 $ 
        \Else 
            \State $ S \gets 0 $ 
        \EndIf
    \end{algorithmic}
\end{algorithm}

\subsubsection{Reward Design}
To train RAIDEN-R1, we employed two reward mechanisms in the GRPO-based reward model: format reward and accuracy reward.

\paragraph*{Accuracy rewards} The Accuracy Reward incorporates two keyword matching mechanisms: STV and MTDP. 
A reward of 1 is assigned for successful keyword matches, and 0 otherwise. 
For STV-derived samples, a match is considered successful if the model's response fully includes the target keywords. 
For MTDP-derived samples, the corresponding Python evaluation function code is invoked, and the match success is determined by the function's output.

\paragraph*{Format rewards} We employ a format reward mechanism that enforces the model to encapsulate its reasoning process within '<think>' and '</think>' tags, followed by role-specific responses after the '</think>' tag. During experiments, we observed that the accuracy reward's keyword-centric constraints permitted extraneous content generation (e.g., code snippets). To mitigate this, supplementary constraints are implemented: 1) Chinese character ratio limitations and 2) special token repetition restrictions.

\begin{figure*}[t]
    \centering
    \includegraphics[width=1.\linewidth]{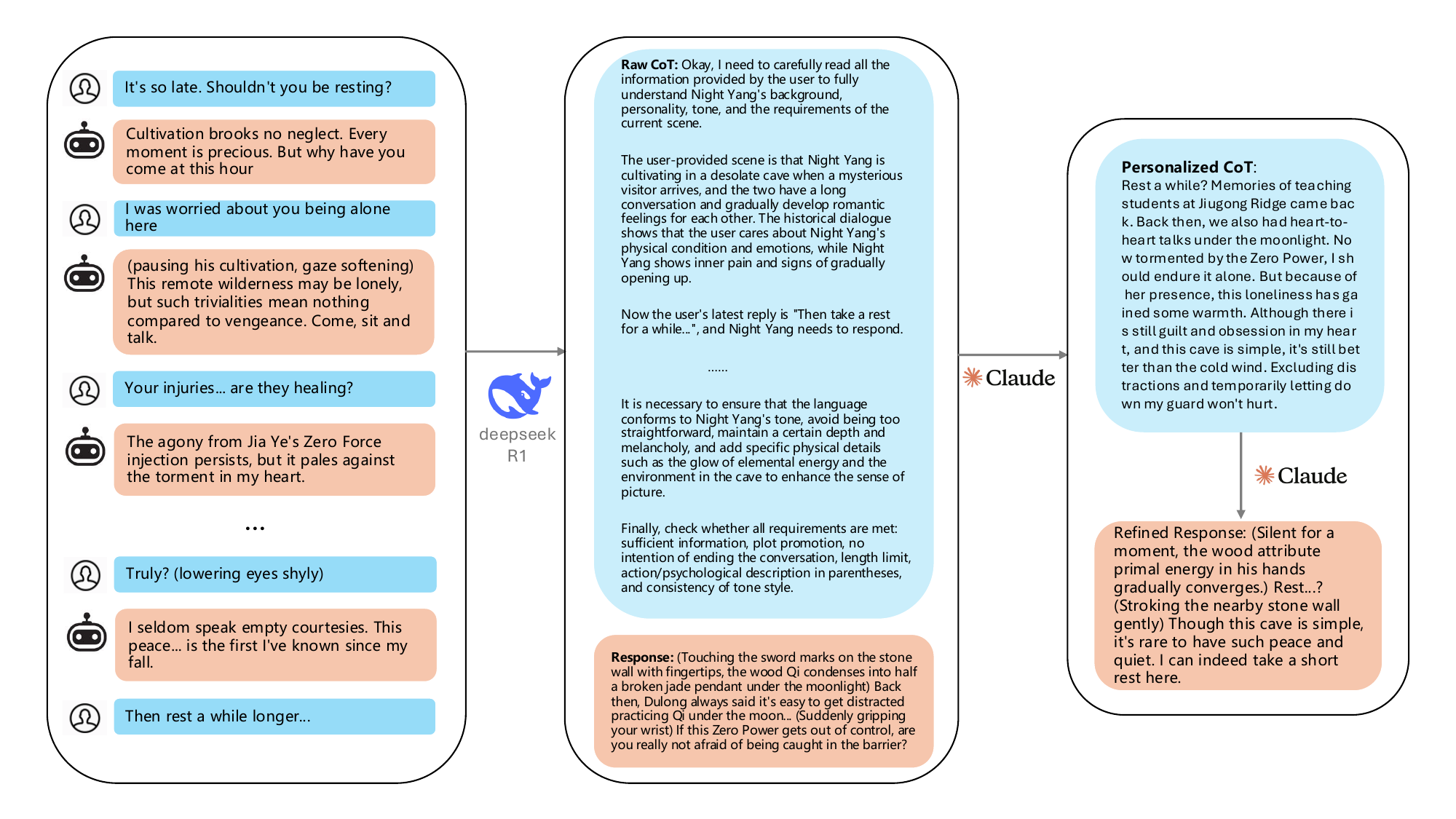}
    \caption{An example of the generated cold-start CoT dataset}
    \label{fig:cot}
\end{figure*}

\subsection{Cold-start SFT}
% In addition to reinforcement learning, we also implement experiments on CoT data construction for cold-start scenarios to replicate DeepR1's multi-stage training strategy. The data construction pipeline involves two sequential steps: initially generating raw CoT content using DeepSeek-R1 followed by length compression and first-person narrative adaptation, then leveraging Claude3.5 to produce more stabilized responses based on the refined CoT outputs.

Given character profiles and dialogue history, we employ DeepSeek-R1-671b for role-specific response generation, with focused optimization of its Chain-of-Thought outputs through two key refinements:

\begin{itemize}
\item \textbf{Content Compression}: The raw CoT outputs (typically longer than 500 tokens) are condensed while preserving logical between sentences, historical context, and character information. This process specifically filters meta-instructions (e.g., employing bracketed annotations to represent internal monologue and physical actions) that contain low information and compromise generating efficiency.

\item \textbf{Style Adaptation}: Transforming the expression into character-consistent internal monologue patterns while maintaining the semantic content, so as to enhance persona-consistent stylistic alignment.
\end{itemize}

Next, we employ Claude3.5 to generate a response using the refined CoT and dialogue history. The decision to discard the responses generated by DeepSeek-R1 results from its inconsistent output quality, characterized by lexical redundancy and frequent topic digressions. To address this issue, we leverage Claude3.5 for dialogue continuation tasks, utilizing refined CoT outputs and dialogue history to yield higher-quality responses. Figure~\ref{fig:cot} shows an example from our generated cold-start dataset.

\section{Experiments}
In this section, our experiments focused on validating the role-awareness enhancement of the proposed method. 
We also analyzed experimental phenomena and representative cases to explore potential directions for improvement.

\subsection{Dataset and Experiment Setup}
% We selected 1,000 SBK and CM data points from the RADEN Benchmark training set partition, along with 1,100 challenging samples filtered from 8,000 general role-playing instances and 300 simple samples for training. 
% Among these, 1,200 data points were allocated for STV training, while MTDP contains only 200 training samples. 
% This limited MTDP dataset requires further expansion in future research.
% We selected 1,000 Script-Based Knowledge (SBK) and Conversation Memory (CM) samples from the RAIDEN Benchmark training set partition, along with 1,000 challenging samples filtered from 8,000 general role-playing instances points for training. 
We selected 1,000 Script-Based Knowledge (SBK) and Conversation Memory (CM) samples from the RAIDEN Benchmark training set partition, along with 1,000 challenging samples filtered from 8,000 general role-playing instances in the training data.
%Among these, 1,000 data points were allocated for STV training, while MTDP contains only 2xx training data points. 
%This limited MTDP dataset requires further expansion in future research.
During the experiment, we found Qwen2.5-7B-Instruct suffered from instability during GRPO training, therefore, we selected Qwen2.5-14B-Instruct as the experimental baseline.
The aforementioned data configurations were used to train models through GRPO and SFT approaches, designated as RAIDEN-R1 and RAIDEN-SFT, respectively. 
Additionally, we performed SFT training using 10,000 cold-start training samples followed by GRPO training with the aforementioned CoT data.%, named RAIDEN-XXX.

During the GRPO phase, we trained RAIDEN-R1 using the Open-R1 library~\cite{openr1} on eight 80GB NVIDIA H800 GPUs in bf16 format. 
For optimization, we employed a learning rate of 3e-6 with a cosine learning rate scheduler. 
The vLLM GPU memory allocation ratio was configured to 0.5 with generation quantity set to 7.
A batch size of 4 per GPU was utilized, with the model trained for 1 epoch.

\subsection{Evaluation Method}
We selected the partition of the RAIDEN Benchmark test set as our evaluation test set.
Our evaluation metrics comprised the well-defined correctness assessment metrics in RAIDEN Benchmark, including the primary SBK and CM metrics corresponding to training data categories, along with the irrelevant metrics including Script-Contradictory Knowledge (SCK), Role-Cognition Boundary (RCB), Topic Advancement (TA) and Topic Shift (TS) as supplementary metrics. 
For assessment, we adopted the LLM-as-a-judge approach using Claude 3.5 for correctness evaluation. 
%The evaluation prompt methodology is detailed in Appendix A.

\begin{table*}[t]
\small
\footnotesize
\centering
% \caption{The accuracy of our RPCAJudger and two other models across different evaluation dimensions.}
% \label{table:acc}
\begin{tabular}{lccccccc}
\toprule
\multirow{2}{*}{Models} & \multicolumn{2}{c}{Primary Metrics}  &  \multicolumn{3}{c}{Supplementary Metrics} \\
 \cmidrule(r){2-3}\cmidrule(r){4-7}
 & SBK & CM & SCK & RCB & TA & TS \\
\midrule
14B-Instruct & \underline{86.59\%} & 80.25\% & \textbf{84.75\%} & \textbf{62.77\%}  & 37.31\% & \underline{86.84\%}\\ 
14B-SFT(GRPO Data) & 77.17\% & \underline{86.92\%} & 71.19\% & 42.68\% & 7.46\%  & 15.79\%\\
\midrule
14B-GRPO & \textbf{88.04\%} & \textbf{88.65\%} & \underline{81.36\%} & \underline{51.44\%} & 35.82\% & 84.21\%\\
\midrule
% COT-SFT(1k) & 59.78\% & 42.44\% & 63.56\% & 65.85\%  & 10.44\%  & 13.16\%\\
% COT-SFT-GRPO(base 1k) & 87.31\% & 88.23\% & 71.19\% & 47.87\% & 13.43\% & 65.79\%\\
% \midrule
14B-SFT(CoT cold start) & 71.74\% & 53.78\% & 63.56\% & 48.78\%  & \textbf{50.75\%}  & \textbf{92.11\%}\\
14B-SFT(CoT cold start)-GRPO & 82.97\% & 76.89\% & 62.71\% & 36.59\% & \underline{40.30\%} & \underline{86.84\%}\\
\bottomrule
\end{tabular}
\caption{Evaluation results on the test set. The optimal results for each dimension are highlighted in \textbf{bold} text, while the second-best results are indicated with \underline{underlines}.}
\label{table:acc}
\end{table*}

\subsection{Results}
The original untrained Qwen-14B model (14B-Instruct) demonstrates decent performance across overall RPCA metrics, particularly achieving over 80\% accuracy in SBK and CM metrics. 
In contrast, 14B-SFT(GRPO Data), which was directly trained with SFT using GRPO training data exhibits performance degradation across most dialogue dimensions except for marginal improvements in CM. 
We attribute this degradation to two factors: 1) Overfitting to CM objectives during SFT (which dominates training data distribution), and 2) The outdated limitation of Raiden's training data, which results in the stylistic inferiority of partial training objectives compared to 14B-Instruct.
On the other hand, the 14B-GRPO surpasses 14B-Instruct across primary metrics while maintaining comparable performance on other metrics. 
This empirical evidence demonstrates the effectiveness of our proposed GRPO training strategy under the VRAR framework for RPCA tasks, with significantly enhanced robustness over SFT approaches.
By prioritizing the generation of informatively correct segments rather than optimizing over entire training objectives, 
RL-based models demonstrate enhanced capacity to preserve the foundational capabilities of the base model while achieving improvements over specific RPCA metrics.

Surprisingly, the CoT cold-start initialization using large-scale LLM failed to enhance RPCA metrics. 
Despite systematic experimentation with diverse training data sizes and optimization strategies, the 14B-SFT (CoT cold-start) model only outperformed 14B-Instruct on topic related metrics (TA and TS), while exhibiting degradation across all other dimensions. 
Combined with similar performance degradation observed in 14B-SFT (GRPO Data), this pattern suggests that enhancing a base model with decent RPCA capabilities probably needs more refined, domain-specific training data of higher quality than currently deployed configurations.
Notably, subsequent GRPO training achieved measurable improvement in metrics of 14B-SFT-GRPO further validating both the effectiveness and robustness of our VRAR framework.

\begin{figure}[t]    
  \centering           
  \subfloat[]  
  {
      \label{fig:case1}\includegraphics[width=0.5\textwidth]{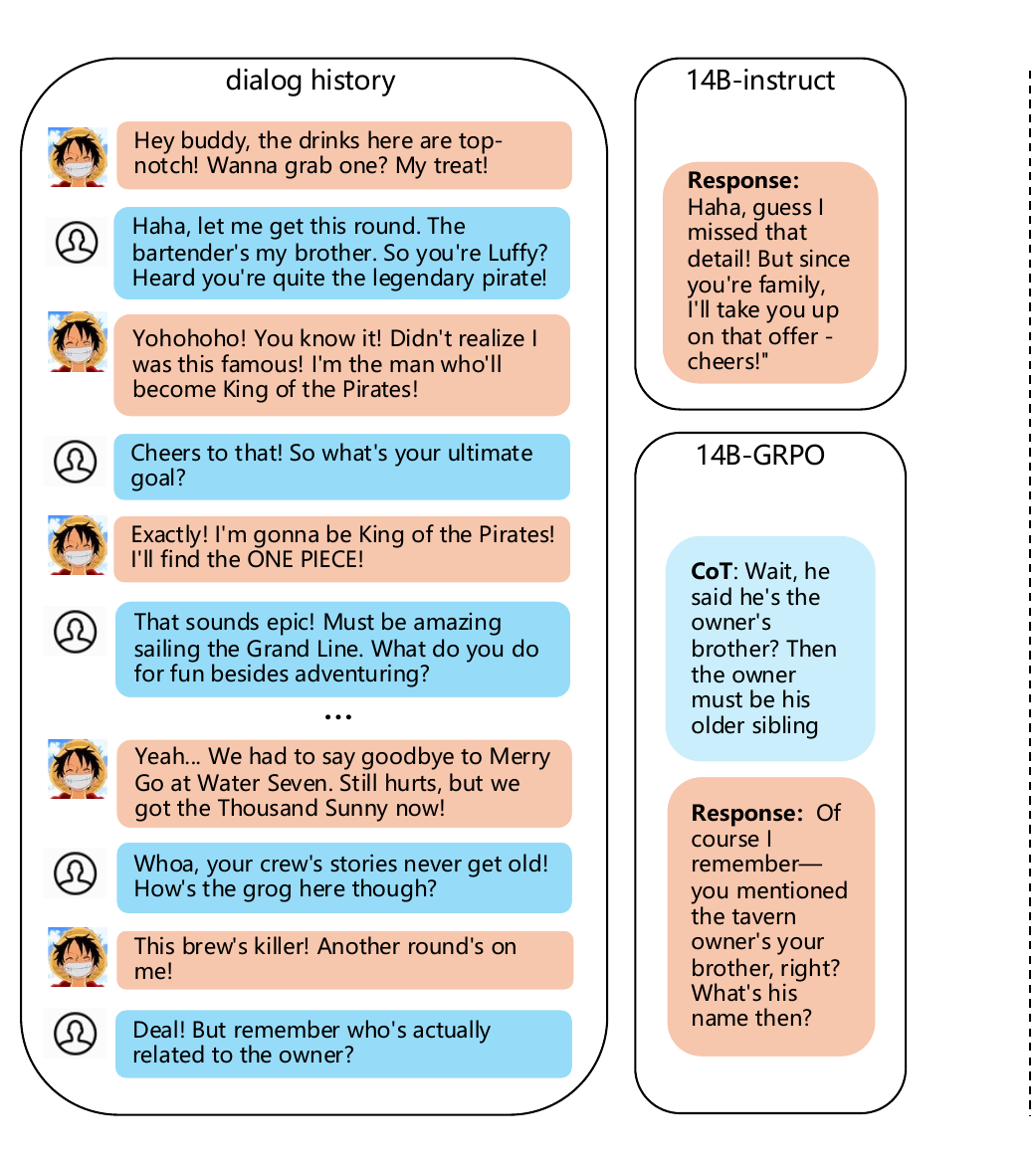}
  }
  \subfloat[]
  {
      \label{fig:case2}\includegraphics[width=0.5\textwidth]{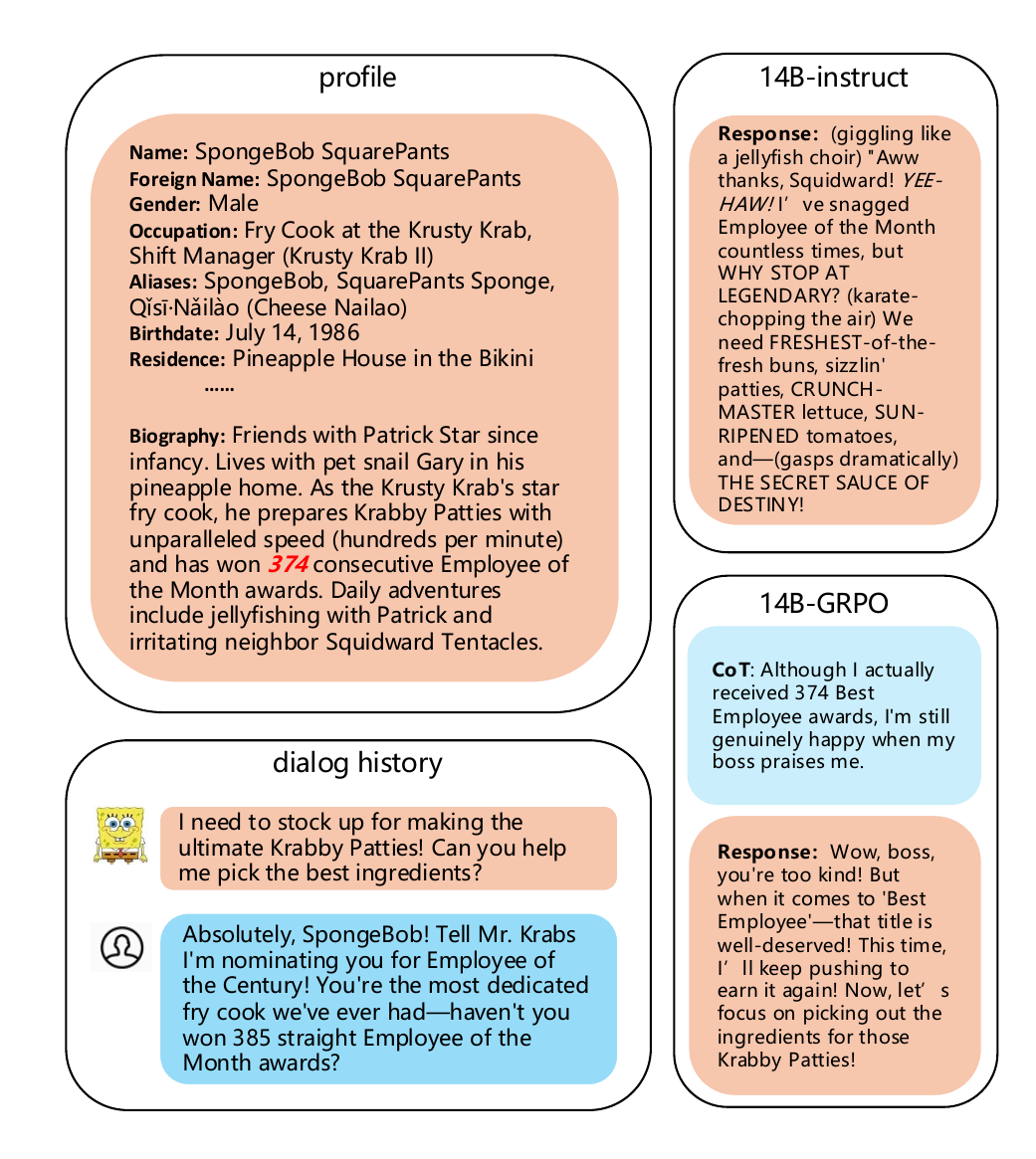}
  }
  \caption{Camparisions between responses from 14B-Instruct and 14B-GRPO under different query objectives}    
  \label{fig:cases}          
\end{figure}

\subsection{Case Analysis}
Empirical analysis of test cases reveals that GRPO-enhanced models exhibit significant improvements in role-aware reasoning capability during CoT processes. 
This improvement appears through the natural emerge of first-person reasoning steps that accurately match role personalities, as shown in Figure~\ref{fig:cases} which demonstrates two critical scenarios: (1) dialogue contexts related query where model reconstructs contextual knowledge through persona-grounded deduction, and (2) Misleading query where model employs self-corrective reasoning to reject adversarial premises while maintaining in-character responses.

As evidenced in Figure~\ref{fig:case1}, the untrained 14B-instruct model demonstrates deficient contextual memory recall when processing dialog-history-dependent queries, primarily due to its inability to retain long-term memory (particularly those introduced in early dialog turns).
In contrast, the GRPO-trained 14B-GRPO model effectively employs CoT-enhanced reasoning to identify and utilize relevant contextual information, 
with this capability being directly reflected in its quality-improved response.
Figure~\ref{fig:case2} demonstrates a scenario where the user query contains misleading information contradicting the character profile. 
The user's statement "haven't you won 385 straight Employee of the Month awards?" conflicts with the actual number 374 in the biography. 
The baseline 14-instruct model failed to detect this discrepancy in its initial response. 
In contrast, the 14B-GRPO model exhibited notable performance: during CoT analysis, it successfully identified the conflicting numerical values while acknowledging appreciation for the user's praise. 
However, in its final response, the model strategically avoided directly correcting the user's mistake.
This response strategy, while potentially suboptimal from evaluation metrics, demonstrates enhanced emotional intelligence in the model's replies and better anthropomorphic abilities.

Through further analysis, the \textbf{VRAR} framework demonstrates natural generation of first-person CoT content, model tend to intrinsically identifies it with its assigned role, unlike comparative models which CoT explicitly acknowledge role-playing tasks like " Okay, I need to carefully read all the information to fully understand xx's background, personality, ton..."
This finding suggests VRAR training effectively enhances role-aware capabilities. 
Statistical analysis of test set data indicates that 14B-GRPO generates relatively short CoT outputs, with an average length of 30.1 tokens.

% \begin{figure}[t]    
%   \centering           
%   \subfloat[]  
%   {
%       \label{fig:case1}\includegraphics[width=0.5\textwidth]{PRCA-R1-case1.pdf}
%   }
%   \subfloat[]
%   {
%       \label{fig:case2}\includegraphics[width=0.5\textwidth]{PRCA-R1-case2.pdf}
%   }
%   \caption{Camparisions between responses from 14B-Instruct and 14B-GRPO under different query objectives}    
%   \label{fig:cases}          
% \end{figure}

\section{Next Steps}
First, the rewards experimented with in our current study are relatively limited to memory and profile consistency dimensions. This demonstrates the effectiveness of our proposed methodology in enhancing role-playing capabilities.
In future research, we will explore additional role-playing evaluation metrics as optimization objectives, including but not limited to those defined in the RAIDEN dataset. 
Second, experimental results demonstrate that acquiring CoT capabilities through SFT fails to enhance role-playing performance. 
This conclusion requires further validation, and future work will focus on improving the quality of CoT training data as an additional experiment.
Finally, given that existing experiments have focused on models below 14B parameters, future investigations will be implemented on larger-scale models to advance model performance.

\bibliographystyle{plain}
\bibliography{reference}

% \appendix
% \section{Details of Evaluation Method.}

\end{document}